\documentclass[sigconf]{acmart}
\usepackage{booktabs} 
\usepackage[caption=false,font=footnotesize]{subfig}
\setcopyright{acmcopyright}

\acmConference[GECCO '17]{the Genetic and Evolutionary Computation Conference 2017}{July 15--19, 2017}{Berlin, Germany}
\acmYear{2017}
\copyrightyear{2017}

\acmPrice{15.00}

\begin{document}
\title{On Self-Adaptive Mutation Restarts for Evolutionary Robotics with Real Rotorcraft}

\author{Gerard David Howard}
\orcid{1234-5678-9012}
\affiliation{%
  \institution{CSIRO}
  \streetaddress{1 Technology Court, Pullenvale}
  \city{Brisbane} 
  \state{Queensland} 
  \postcode{4068}
}
\email{david.howard@csiro.au}

\begin{abstract}
Self-adaptive parameters are increasingly used in the field of Evolutionary Robotics, as they allow key evolutionary rates to vary autonomously in a context-sensitive manner throughout the optimisation process.  A significant limitation to self-adaptive mutation is that rates can be set unfavourably, which hinders convergence.  Rate restarts are typically employed to remedy this, but thus far have only been applied in Evolutionary Robotics for mutation-only algorithms. This paper focuses on the level at which evolutionary rate restarts are applied in population-based algorithms with $>$1 evolutionary operator.  After testing on a real hexacopter hovering task, we conclude that individual-level restarting results in higher fitness solutions without fitness stagnation, and population restarts provide a more stable rate evolution.  Without restarts, experiments can become stuck in suboptimal controller/rate combinations which can be difficult to escape from.
\end{abstract}

\ccsdesc[500]{Theory of computation~Evolutionary algorithms}
\ccsdesc[500]{Computer systems organization~Evolutionary robotics}
\ccsdesc[300]{Computer systems organization~Real-time systems}
\ccsdesc[300]{Computing methodologies~Learning settings}
\ccsdesc[300]{Applied computing~Aerospace}


\keywords{Evolutionary robotics, Self-adaptation, evolvable hardware, differential evolution}

\maketitle

\section{Introduction}

One of the longstanding issues in Evolutionary Robotics (ER) \footnote{The use of Evolutionary Algorithms to create robot controllers and/or morphologies.} research~\cite{nolfi-flor-01} is the assessment of phenotype and reward of a suitable fitness.  There are two generally accepted methods, either (i) simulate the robot together with its environment, or (ii) physically test the robot in the real world.   Simulation (e.g.~\cite{howard-snn-quad}) is a popular choice as it is parallelisable, and, depending on model complexity, may run many times faster than real-time.  Simulation suffers from the 'reality gap'~\cite{jakobi1995noise}, whereby the necessarily-abstracted physical laws present in the simulation inaccurately represent real-world conditions, resulting in performance degredation when the former is transferred to the latter.  Early efforts to combat this effect focused on the application of suitable levels of noise~\cite{jakobi1995noise}; recent research includes selecting controllers that are transferrable,(e.g., simulated performance is close to real performance)~\cite{Koos:2010:CRG:1830483.1830505}, and coevolutionary methods that use real measurements to inform the simulator~\cite{bongard2006resilient}, and as such can be seen as a hybrid of the two approaches.

Conversely, physical testing (e.g.,~\cite{montanier:inria-00566898}) guarantees that the results of the evolution work in reality, capturing dynamics and physical effects that may be missing from a simulator.  In this scenario, optimisation times are long, as evaluations are inherently limited to real-time, and repeatable test environments need to be engineered to ensure fair test conditions.  Additionally, working with real robots raises a number of practical issues, as highlighted by early work~\cite{harvey1994seeing} that approximated a differential-drive robot using a gantry-mounted camera, which was simpler to reset and maintain.  In general, the choice of simulation vs. reality can be framed as a trade-off between evaluation {\em speed} (how quickly we can evolve a controller) and {\em performance} (how well it works in the real world).

The issues of performing ER with real robots are exacerbated when flying robots are considered, as testing stochastically-generated controllers on real flying robots can be destructive.  Recently, the evolution of controllers directly onto real flying robots (specifically the popular and versatile hexacopter Unmanned Aerial Vehicle (UAV)) has been made possible, through a platform that uses a combination of physical tethers and the real-time monitoring and recovery from dangerous states~\cite{howard2015platform}.  The platform safely, repeatedly, and non-destructively evolves controllers for UAVs directly on the robot (i.e., without modelling), which provides the benefits that (i) the controllers are guaranteed to work on the UAV in reality, and (ii) effects of hardware state of the UAV (e.g., propeller wear, payload configurations) on the flight dynamics are implicitly captured.  We can describe this platform as having high {\em performance} but low {\em speed}, and increasing the speed of evolution in this platform is the focus of this research.

As we cannot significantly increase the speed of an individual evaluation without modelling (which we preclude as provision of a UAV model of sufficient fidelity to accurately capture the physical reality of every conceivable payload and hardware state is unrealistic), we instead consider reducing the number of evaluations required.  Self-Adaptive (SA) mutation (e.g.,~\cite{771166}) is a promising approach that has previously been used to reduce the number of generations required to generate high-fitness solutions in simulated ER experiments~\cite{howard2012evolution}, and has shown promise in hardware ER~\cite{5585926}.  SA learning rates (e.g., mutation, crossover) can adapt to the instantaneous requirements of the problem considered in a context-sensitive manner, not only at the start of the experiment but throughout the evolutionary process.  SA is particularly suited to our problem, as the platform will optimise myriad different UAVs and payloads, and as such is likely prefer different learning rates from experiment to experiment.

An issue with SA mutation is fitness stagnation, enacted through a combination of suboptimal learning rates and locally-optimal controllers, which cannot improve as their rates are suboptimal.  In the context of ER, this is especially problematic as any experimental time wasted is real-time.  Rate restarts are shown to be an effective technique to dissuade such behaviour~\cite{montanier:inria-00566898}.  The question is, when population-based EA's are considered, do we restart the mutation rates based on population fitness stagnation, or rather based on individual fitness stagnation?

In this paper we present the results of an experiment that seeks to answer this question.  We test an individual-level restart strategy and a population-restart strategy, comparing to benchmarks of static rates, and self-adaptive rates with no restarts.  The performance of each strategy is assessed on a task where a real hexacopter is optimised for hovering behaviour in presence of a significant wind disturbance.


\section{Background}
In this section we provide research in two relevant areas; Evolutionary Robotics, and Self-Adaptation.

\subsection{ER with Flying Robots} 
Due to the potentially destructive nature of stochastically optimising controllers for flying robots, simulation is popular~\cite{4631139,perhinschi1997modified,1688525,howard-snn-quad,1302409}.  Simulation also allows evaluation to occur faster than real-time, however the faster the simulation is, the more abstracted the underlying model of reality tends to be.  This results in controllers transferred from simulation to reality being unable to cross the 'reality gap', e.g.,~\cite{phillips1996helicopter}.  This is evident even in recent work which evolved behaviour trees to allow a micro UAV to escape from a room by flying through a window~\cite{7412843}, resulting in a simulated escape rate of 88$\%$, which was reduced to 46$\%$ in reality and could only be increased to 54$\%$ through manual rule tweaking.

Attempts to directly evolve control for real flying robots are limited.  A blimp controller is successfully evolved~\cite{wheels-to-wings}, but the slow dynamics of the blimp simplifies recovery from dangerous states.  Control of a miniature helicopter~\cite{4983001} is evolved, although only height and yaw control are optimized.

Coevolutionary methods are applied to force quadrotor models (represented using Genetic Programming trees) to match real recorded flight data in a system-identification approach~\cite{sussex23829}, however the experimentation is focused on modelling rather than controller optimization.

Controllers are evolved on real hexacopters using a Bee Colony Algorithm~\cite{ghiglino2015online}, which is demonstrated to work as both an online and offline optimiser, with only small performance differences between the two modes.  However, state estimation requires an expensive infrared tracking system, and frequent human intervention is required to e.g., change batteries.

Recently, a platform is demonstrated that allows for safe and repeatable 24/7 controller optimisation of any multi-rotor (with certain size limitations)~\cite{howard2015platform}.  As controller are directly evolved, controllers are guaranteed to work on the real robot, accounting for any attached payload and hardware variability.  However, evaluation is limited to real-time.  To improve the efficiency of the platform, self-adaptation is proposed as a method of reducing the number of evaluations required.

\subsection{Self-Adaptation}

Self-adaptation (see, e.g.,~\cite{771166} for an overview) allows key evolutionary parameters to vary throughout the optimization process, allowing suitable rates to be found for an instantaneous evolutionary state.  Due to their real-time limitation, hardware ER experiments typically have a low feasible number of fitness evaluations that can be executed~\cite{5729613}.  SA has been used to reduce the time spent evaluating the population by optimisating the controller evaluation times explicitly, in simulation~\cite{dinu2013self}.  Further research~\cite{5729613} identifies three common parameters that can be varied; population size, mutation rate, and the controller evaluation period, together with a re-evaluation rate which is less commonly but necessary in online scenarios to achieve more reliable fitness estimates.  The authors conclude that mutation rate has the most significant effect in reducing evaluation times.  It is therefore mutation rates that we focus on in this study.

Fitness stagnation is a common problem when using SA, as rates may be set that prevent successful location of the global optimal solution.  We note the effectiveness of rate restarts~\cite{montanier:inria-00566898} in countering the effects of fitness stagnation.   Performance-based restarting of unfavourable rates is shown to (i) dissuade premature convergence into unfavourable areas of the rate space, and (ii) `rescue' the optimisation process from unfavourable rate settings.  As~\cite{montanier:inria-00566898} uses a simple 1+1 Evolution Strategy~\cite{eigen1973ingo}, the authors do not consider the different effects that may be observed if the rate restart is applied on the level of the individual, vs. the level of the population.  As fitness stagnation is still an issue with population-based SA, this question is particularly relevant for our application.

For our purposes, an individual-level restart involves the mutation rates of that individual being reset if its fitness doesn't improve for $n$ consecutive generations.  A population-level restart will restart every individual's mutation rates if none of the individuals can generate a fitness improvement for $n$ consecutive generations.

Conceptually, it is not obvious which would be preferrable --- individual restarts may be too unstable when combined with the self-adaptation of the rates themselves, but the restarts are triggered immediately on an individual.  Conversely, population-level restarts will present a more stable evolutionary process, but the focus on improving global fitness means that individuals who are globally suboptimal, but with good rate settings, may be adversely affected.

We are motivated to investigate the effects of these two restart strategies on the performance of an ER experiment, and intend to produce results that will inform the use of rate restarts by other researchers.

\section{Experimental Platform}
\begin{figure}[t]
\begin{center}
 \subfloat{\includegraphics[width=8.5cm]{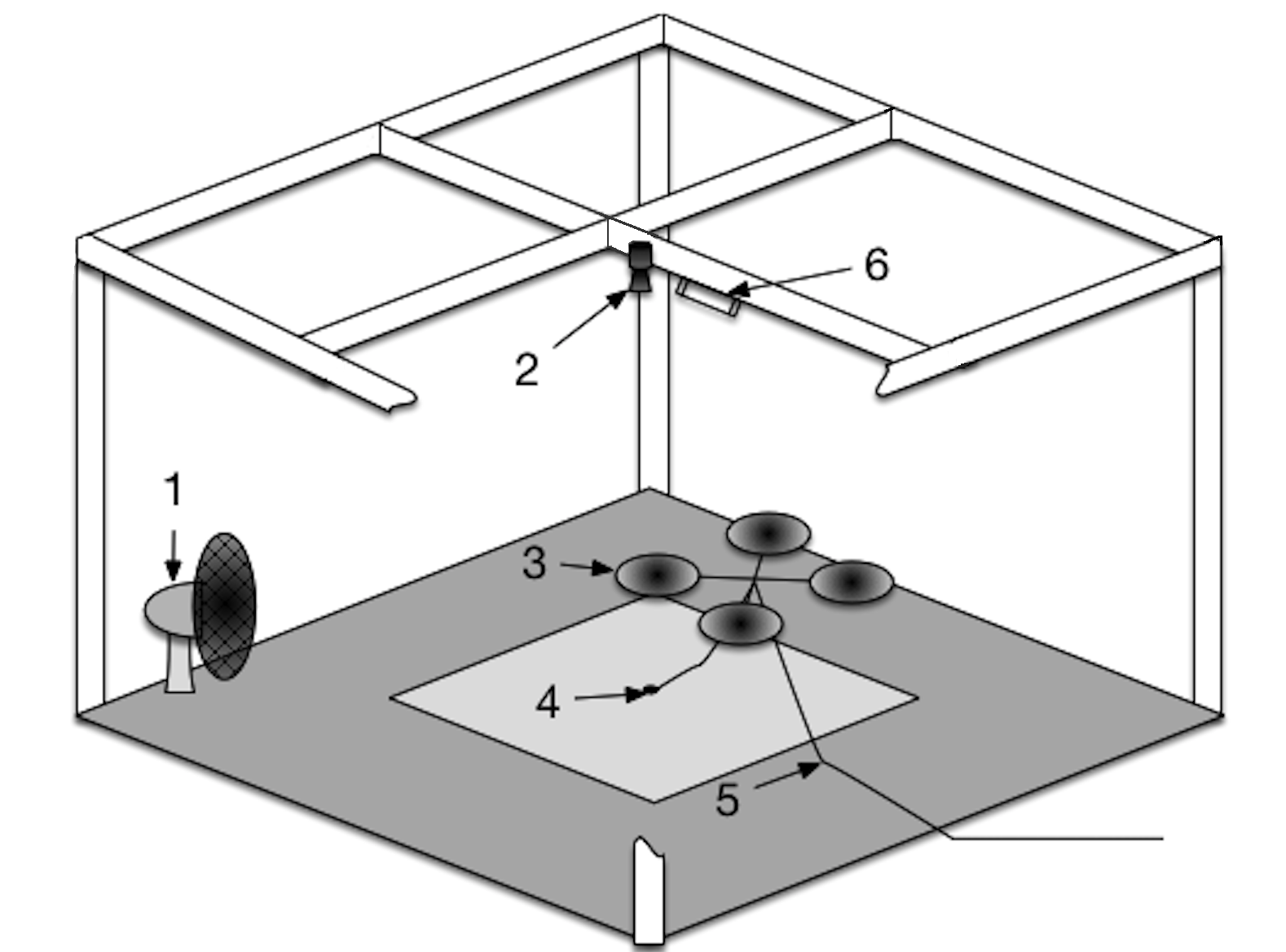} } 
\end{center}
\caption[]{The platform, showing (1) the fan, (2) camera, (3) hexacopter, (4) physical tether, (5) data/power tether, and (6) light.  The camera height is 200cm and padded floor area is 271cm$^2$.  The light grey floor area depicts a standard flight area of $\approx$60$cm$ in $x$ and $y$, and 20cm in $z$.}
\label{setup_new}
\end{figure}

Experimentation occurs on our optimisation platform.  We refer the interested reader to our previous research~\cite{howard2015platform} for a full algorithmic description, as well as a similar platform for multi-legged robots~\cite{heijnen2017ICRA}.  Briefly, the platform comprises a solid floor which is covered with foam matting.  The hexacopter is anchored to the floor with nylon wires, so that flipping (tilt angles $>60^o$), and excessive rotation ($\pm160^o$) are physically prohibited.  An LED strip light and camera are mounted atop a mesh-covered metal frame, which stands over the floor.  An oscillating fan provides wind disturbances of $\approx$5m/s, with an oscillation period of 10 seconds and total traversal angle of $120^o$.  A 24V cable provides constant power, and a serial cable connects to the host PC, which manages and monitors experiments using the real-time Extended State Machine (ESM) framework~\cite{merz2006control}. See Fig.\ref{setup_new}.  

\section{Controllers}
The platform evolves a population of hexacopter controllers.  We use Proportional-Integral-Derivative (PID) controllers\cite{PIDbook} as they are a {\em de facto} representation, and compatible with most commercially-available flight controllers, which increases the generality of the platform.  PIDs have previously been shown to be amenable to evolutionary optimisation --- see\cite{fleming2002evolutionary} for a survey.

\begin{figure*}[ht]
\begin{center}
 \subfloat{\includegraphics[width=16cm]{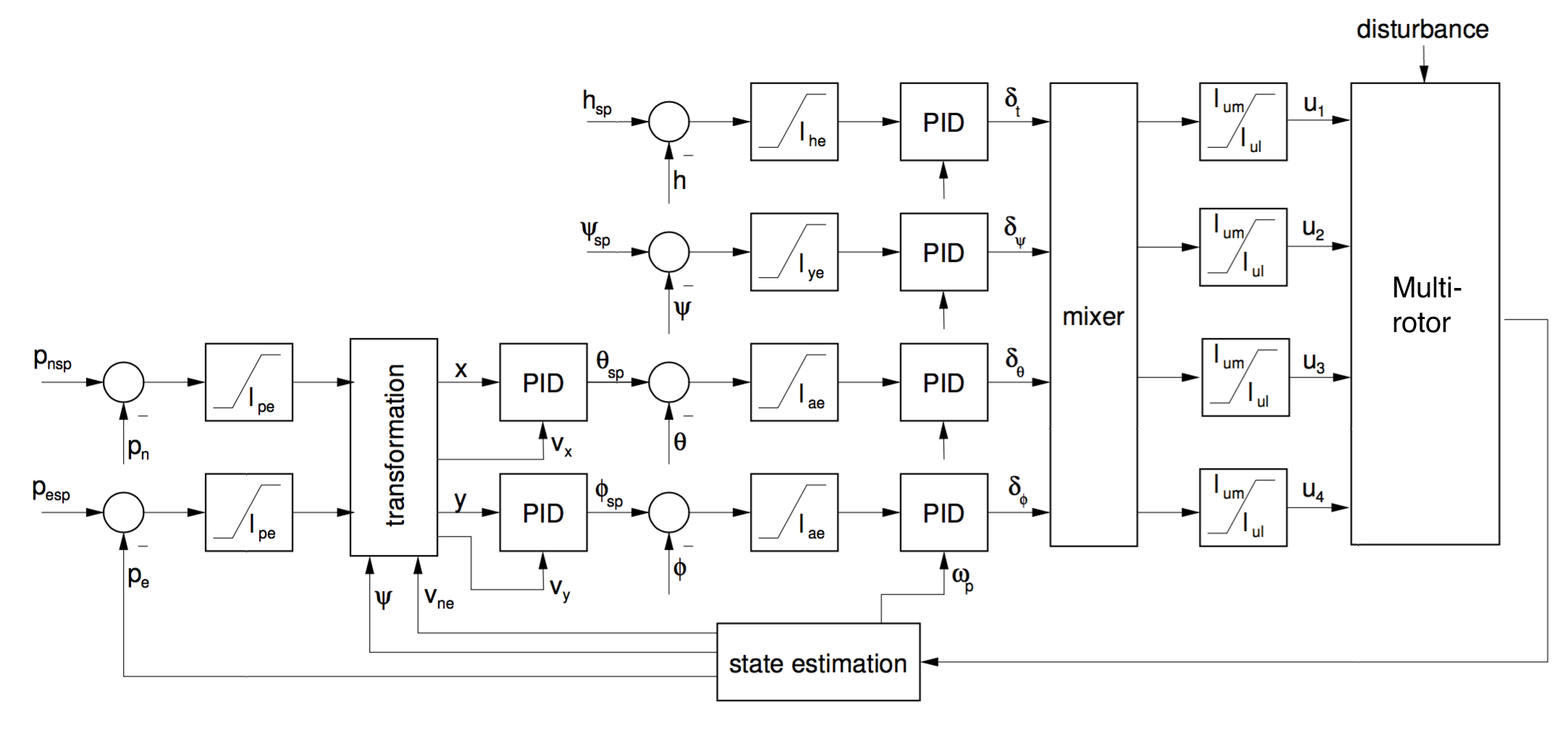} }
\end{center}
\caption[]{PID control structure, showing attitude and position loops.  Parameters $l_{he}$ $l_{ye}$ $l_{ae}$ denote error limits for height yaw and attitude respectively. $l_{ul}$ and $l_{um}$ are minimum and maximum motor commands, and $\delta_{\phi}$, $\delta_{\theta}$, $\delta_{\psi}$, and $\delta_{t}$ are command inputs to a mixer which outputs speed controller commands $u_1$, $u_2$, $u_3$, and $u_4$.  The disturbance is input from the fan.}
\label{PID-structure}
\end{figure*}

A two-loop PID structure controls the hexacopter's position and attitude; see Fig.\ref{PID-structure}.  Horizontal position ($p_n$ and $p_e$) is controlled by the outer loop, and attitude ($\phi$, $\theta$, $\psi$) and height $h$ by the inner loop.  The outer-loop PIDs generate setpoints $\theta_{sp}$ and $\psi_{sp}$.  Outputs $\delta_{\phi}$, $\delta_{\theta}$, $\delta_{\psi}$, and $\delta_{t}$ represent commanded changes in attitude and thrust, which are scaled in the range of attainable motor PWM signals $l_{ul}$=1000 and $l_{um}$=2000, and passed to a linear mixer which produces one controller command per motor $m$, e.g., $u_1$ to $u_m$.

PID control minimises the error $e$ between the hexacopter's estimated position and attitude, and the current waypoint, following (\ref{pid-eq}).  There are 6 PIDS in all, as the waypoint is represented by a 6-tuple of setpoints for attitude ($\phi_\mathrm{sp}$, $\theta_\mathrm{sp}$, $\psi_\mathrm{sp}$), and position ($p_\mathrm{nsp}$, $p_\mathrm{esp}$, $h_\mathrm{sp}$).  Each variable is limited to a maximum error $l_\mathrm{er}$ (10cm for $h$, 15$^o$ for attitude, 15cm for $p_n/p_e$) before being input to the PID.

Here, $o$ is the PID output, $t$ is the instantaneous time, $\tau$ is the integration timestep from 0 to $t$, and $K_p$, $K_i$, and $K_d$ are controller parameters that define the response of the controller to raw error, integral error, and the derivative of the error respectively.  With three gains per PID and six PIDs, a controller is represented by 18 reals.

\begin{equation}
o(t) = K_pe(t) + K_i \int_0^t e(\tau)d\tau + Kd\dfrac{d}{dt}e(t)
\label{pid-eq}
\end{equation}

\section{Evolutionary Algorithm}
Controllers are evolved using a self-adaptive Differential Evolution (DE).  Specifically, we use DE/rand/1/bin as it has shown promising results in evolving PID gains~\cite{6217801, biswas2009design}.  Per generation, a donor vector $v$ is created for each `parent' individual $p$ as in (\ref{de-eq}), where $F$, $(0< F \leq2)$ is the differential weight, and $r1$, $r2$, and $r3$ are unique individuals that are selected uniform-randomly.

\begin{equation}
{v} = {r3} + F ({r1} - {r2})
\label{de-eq}
\end{equation}

A `child' vector ${c}$ is created by probabilistically replacing elements of ${p}$ with elemnts of ${v}$.  For each vector index $i$, $c_i$ = $v_i$ if $i==R$ or $rand$ $<$ {\em CR}, otherwise $c_i$ = $p_i$. $rand$ is a uniform-random number between 0 and 1, {\em CR}, $(0< CR \leq1)$, is the crossover rate, and $R$ is a random vector index, ensuring $c$ $\neq$ $p$.  The children are evaluated and assigned a fitness $f$, with $c$ replacing its parent $p$ if $f_{c}$ is superior to $f_{p}$.  When every child has been evaluated, the next generation begins.

Self-adaptation is based on an Evolution Strategy, following e.g.,~\cite{eigen1973ingo}, to allow more straightforward comparisons to previous work with evolution strategy operators\cite,{montanier:inria-00566898}.  New population members random-uniformly initialise their {\em CR} and {\em F}, respecting bounds. Child individuals copy their parent's {\em CR} and {\em F}, and modifies them following (\ref{sa_eq}), respecting bounds.  The comparative static baseline rates are {\em CR}=0.5, and {\em F}=0.8, following a brief parameter sweep~\cite{howard2015platform}.

\begin {equation}
\mu \leftarrow \mu * e^{N(0,1)}
\label{sa_eq}
\end {equation}

\subsection{Restart Strategies}

An individual is represented by a controller, plus its fitness $f$, rates {\em CR} and {\em F}, and a restart counter $r$, which is initially 0.  For individual-level restarts, $r$ is incremented for a parent when it's child does not replace it.  For population-level restarts, a global $r$ is incremented for each consecutive generation the best population fitness is not improved.  Global restarts emphasise global fitness improvements by instantaneously restarting all rates at the same time, whereas individual-level restarts encourage each classifier to improve itself without consideration of global population performance.  

 A restart is triggered when $r$==5 \footnote{Experimentally determined to balance stability with response time.  Similar results were noted during an initial parameter search using $r$ = 3, 5, 7, and 9}.  Individual-level restarts affect only the individual's rates, whereas population-level and periodic restarts simultaneously affect every member of the population.  Restarts reinitialise {\em CR} and {\em F} uniform-randomly within their respective ranges, and also resets $r$ to 0.  

 Note that this is a fitness-based restart, as opposed to a rate magnitude-based restart~\cite{montanier:inria-00566898}, where the mutation step-size alone triggers a restart.  As we use two rates, we consider the overall effect of both rates, which can be neatly captured through the ability of an individual (population) to consistently improve it's fitness over consecutive generations.


\section{Test Problem}

Performance is evaluated on a wind-affected hover scenario, with a total evaluation length of 40s. A hexacopter attempts to follow a series of five waypoints; the target waypoint changes deterministically every 8s.  The waypoints are designed to sufficiently excite all of the hexacopter's six degrees of freedom, see Fig.\ref{trajectory}.

\begin{figure}[t!]
\begin{center}
 \subfloat{\includegraphics[width=8.5cm, height=6cm]{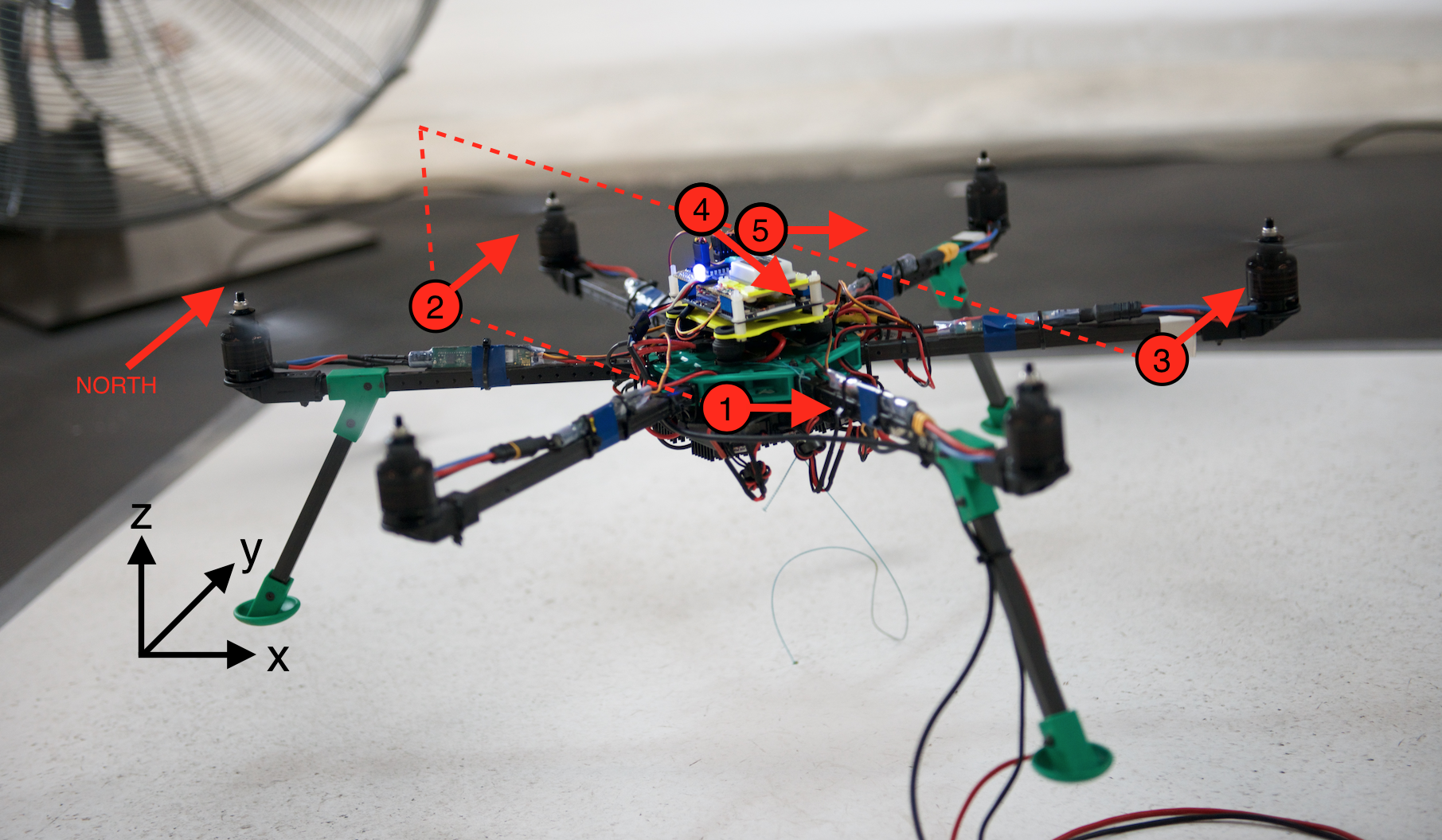} }
\end{center}
\caption[]{The trajectory flown by the hexacopter.  Waypoints change every 8s and are: (1) hover at a height of 10cm with a yaw of 40$^o$, (2) move 8cm North and 8cm West with a yaw of 0$^o$, (3) increase height to 14cm and move 16cm South and 16cm East, (4) return to the centre of the cage with a yaw of 80$^o$, (5) alter yaw to 40$^o$.  The fan can be seen in the top-left of the image.}
\label{trajectory}
\end{figure}

\subsection{Initialisation}
At the start of an experiment, controllers are randomly generated, briefly evaluated, and added to the population if they allow the hexacopter to stay in the air for $>0.2$s.  When the population size reaches $N=20$, the first generation begins.  Initial control parameter ranges are calculated using (\ref{init-eq}), where $l_\mathrm{cmd}$ is a generalised maximum possible command (PWM) for each of the control parameters $V$: $l_\mathrm{cmd}$ for $\phi$/$\theta$/$\psi/h$=500, and $p_n$/$p_e$=15cm.

\begin{equation}
K_\mathrm{pV}=K_\mathrm{iV}=K_\mathrm{dV}=(0,\frac{l_\mathrm{cmd}}{l_\mathrm{er}}]
\label{init-eq}
\end{equation}

\subsection{State Estimation}
The hexacopter's state is estimated at 400Hz.  The hexacopters full state vector comprises: attitude Euler angles (roll $\phi$, pitch $\theta$, yaw $\psi$, at 400Hz), plus angular rates ($\omega_p$, $\omega_q$, $\omega_r$, at 400Hz), and height $h$, at 20Hz, together with position for North $p_n$ and East $p_e$, and velocities $v_n$, $v_e$, $v_h$ (all at 60Hz).  Range limits are provided in Appendix A.

An Inertial measurement unit calculates Euler angles and height (together with a frame-mounted ultrasonic rangefinder).  Attitude angles are processed through a Kalman Filter, and height through a complimentary filter.    Position is measured through a machine vision camera.  Angular rates are derived from two consecutive Euler angles, and velocities calculated through a linear regression of five consecutive position estimates.  This provides a 3D position error $<$5mm and heading error $<$2$^o$.  Position and attitude are used by the controller.   The full state estimate is used to assign fitness and perform health monitoring.

\subsection{Fitness Assignment}

During an evaluation, fitness accumulates at 400Hz by adding a per-Hz fitness measure $f_{cycle}$ (max. 10) to a running total $f$ (max. 160,000).  The composition of $f_{cycle}$ is depicted in Appendix B.  In brief, a high $f$ corresponds to the hexacopter's state closely matching the position and attitude setpoints of the current waypoint.

To account for noisy fitness assessments brought about by imperfect sensors, any controller that completes the full 40s evaluation is immediately reevaluated and assigned the mean fitness.  If the controller completes both evaluations, it is said to be a success.  Successful controllers are reset to their start positions (centre of the floor area with $\psi$=40$^o$) to ensure a fair test between controllers; before this point we are more interested in discovering controllers that can fly rather than accurately comparing controller performance within the population.

\subsection{Health Monitoring}

ESM monitors dangerous hexacopter states, and safely terminates an evaluation if any of the following are detected: $h>$18cm, $v_n$/$v_e>$ 50cm/s, $v_h>$25cm/s, $\phi$/$\theta>$ 15$^o$, maximum yaw error of 45$^o$ exceeded, maximum current draw of 15A exceeded, or maximum rate for upper PWM limit of 75$\frac{1}{s}$ exceeded.  As well as dangerous states, termination also occurs if the hexacopter doesn't move during the first 5s of an evaluation (a time-saving measure), or if the hexacopter lands during an evaluation (touches the ground for $>$1s) having previously been flying.  If a flight is terminated, the controller is assigned its current accumulated fitness.

\section{Experimental Setup}

In its current configuration, the platform executes ten experimental repeats for each of the four restart strategies.  Each repeat optimises 20 controllers over a number of generations until convergence.  Each generation involves the creation of 20 new individuals, which are evaluated on the test problem, and potentially replace current population members.  An evaluation involves an individual's control parameters being used by the hexacopter, and culminates with a fitness value being assigned to the individual.  The experiment ends when each controller in the population can fly for the full 40s evaluation period (convergence).   For brevity, we refer to the different strategies as STATIC (static mutation rates), ADAPT (self-adaptive, no restarts), INDIV (individual-level restarts) and GLOBAL (global-level restarts).  The Mann-Whitney U-test is used to statistically compare the strategies.

\begin{table*}
  \caption{Comparing common performance metrics between the four restart strategies. Standard deviations are shown in parenthesis.  Symbols indicate the strategy is statistically (p$<$0.05) better (higher, for {\em CR} and {\em F}) than $\ddagger$ = GLOBAL, $\dagger$ = STATIC, * = ADAPT, as measured by a Mann-Whitney U-test at p$<$0.05.}
  \label{stats}
  \begin{tabular}{cccccccc}
    \toprule
    Strategy &Conv.   	           	&High $f$ 		  						&Mean $f$ 					&Low $f$ 							&{\em CR} 	 &{\em F}\\
    \midrule
    INDIV 	& 27.6 (10.6)$\dagger$*	& 127185.2 (2716)$\ddagger$$\dagger$*	&122562.6$\dagger$*	(2973)	&117480.6 (4813)$\ddagger$$\dagger$	&0.413 (0.12)&0.756 (0.32)*\\

    GLOBAL 	& 24.4 (9.0)$\dagger$*	& 124036.7 (3376)						&117455.4 (3927) 			&111919.3 (4684)					&0.536 (0.17)&0.758 (0.28)*\\

    ADAPT 	& 32.4 (59.8)$\dagger$ 	& 124151.1 (1528)						&119902.9 (2021)$\dagger$	&115716.2 (2400)$\dagger$			&0.482 (0.24)&0.289 (0.17)\\

    STATIC 	& 70.5 (16.0)		   	& 124020.7 (1429)						&117483   (2172)			&110795.8 (3272) 					&-			 &-	\\
  \end{tabular}
\end{table*}

\section{Analysis}


{\bf Convergence.} Table~\ref{stats} and Fig\ref{trajectory}(a) reveal that all three of the self-adaptive strategies converge more rapidly than STATIC (all p$<$0.05), showing the benefits of self-adaptation over STATIC (although we note that STATIC is a baseline only).  GLOBAL displays the best mean convergence generation (24.4), which is significantly better than STATIC and ADAPT (p$<$0.05), and similar to INDIV (27.6).  Compared to GLOBAL and INDIV, ADAPT displays two outlier experiments (with convergence generations 171 and 173), resulting in the statistically significant differences between them (p$<$0.05).  We conclude that self-adaptation is beneficial to the evolutionary process, but restarts are required to prevent unsuitable rate settings.

\begin{figure*}[t!]
\begin{center}
 \subfloat{\includegraphics[height=5.9cm]{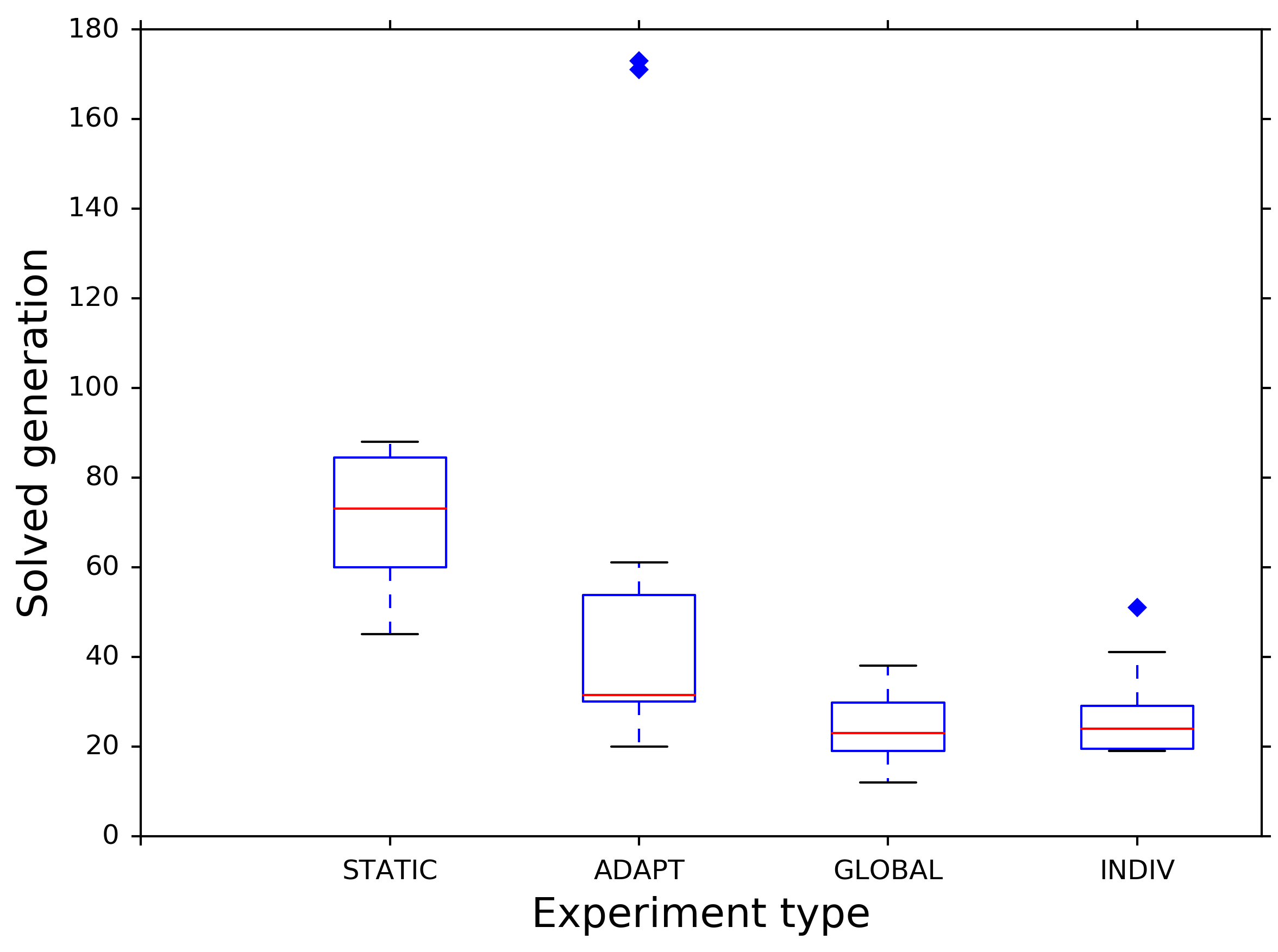} }
 \subfloat{\includegraphics[height=5.9cm]{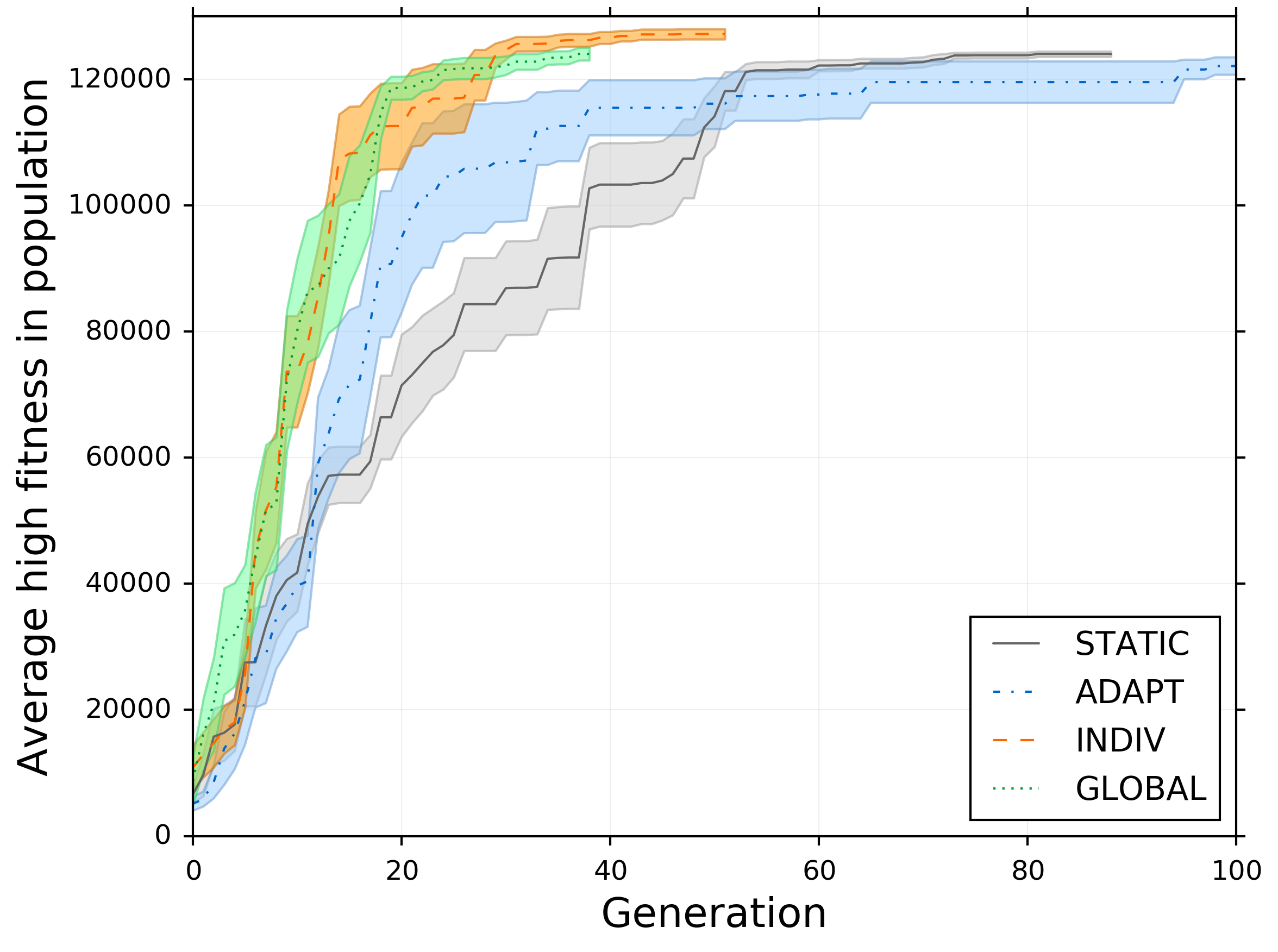} }\\
 \subfloat{\includegraphics[height=5.9cm]{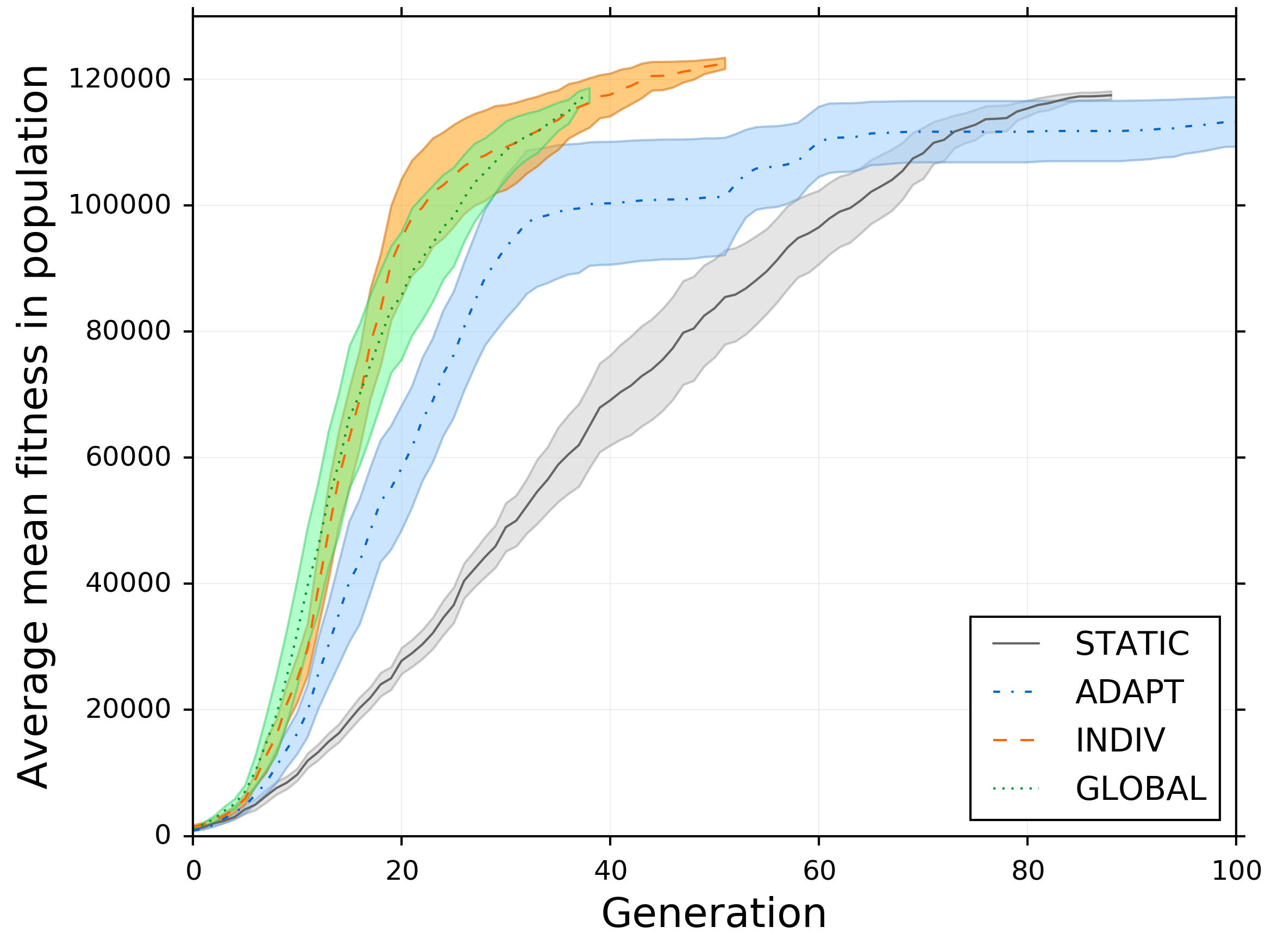} }
 \subfloat{\includegraphics[height=5.9cm]{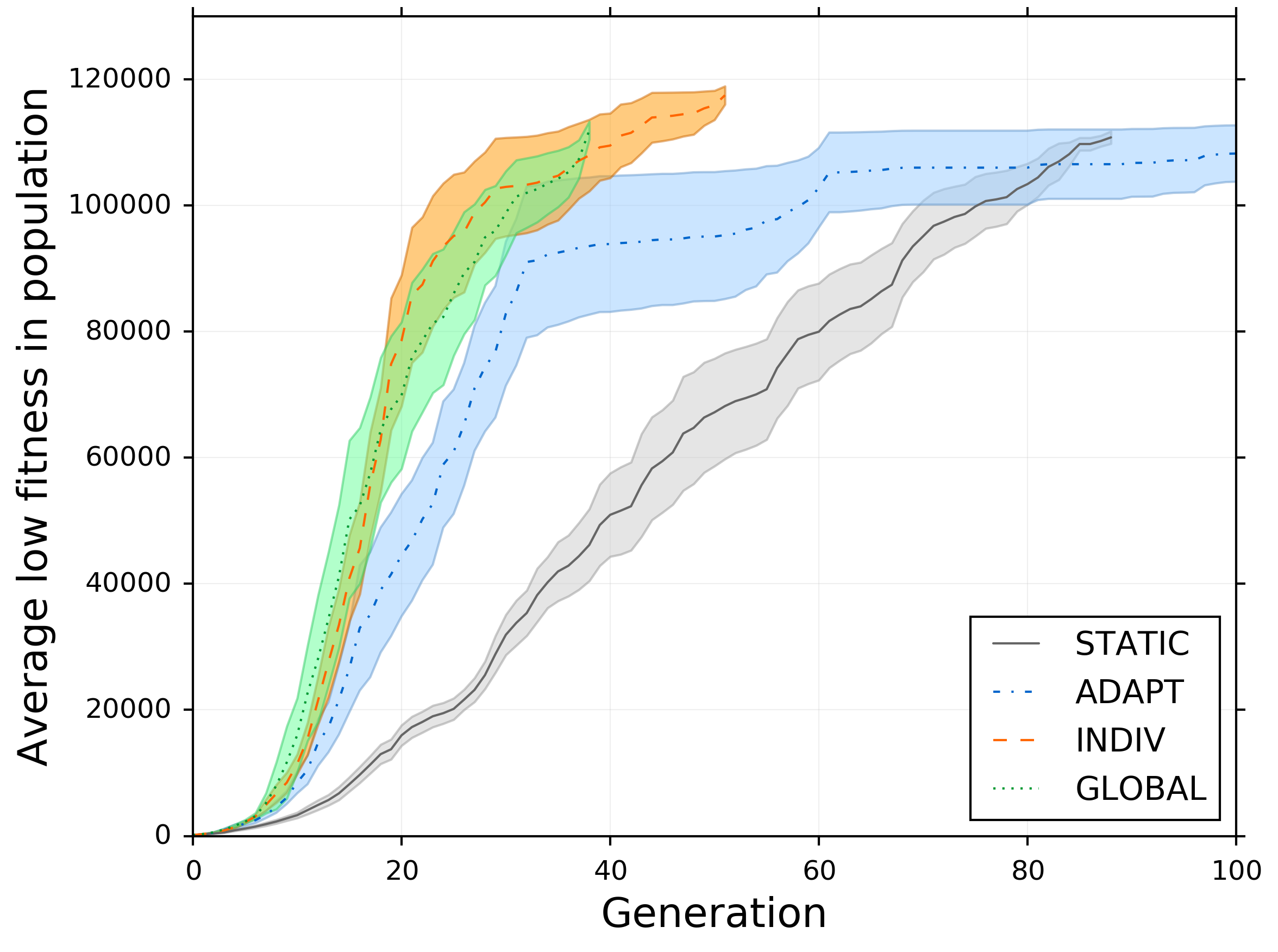} }
\end{center}
\caption[]{(a) A comparative boxplot showing convergence generations, highlighting the outliers for SA (171 and 173).  Outliers are shown with diamonds.(b) High, (c) mean, and (d) low fitness averages for the four strategies over the first 100 generations.  Lines are plotted until all repeats for a given strategy have converged.}
\label{trajectory}
\end{figure*}

{\bf Fitness.}  Fitness trends can be seen in Fig.~\ref{trajectory}(b)-(d)The mean highest fitness for GLOBAL ($f$=124036.7), ADAPT ($f$=124151.1), and STATIC ($f$=124020.7) are statistically similar. INDIV ($f$=127185.2, p$<$0.05) has statistically better high fitness than all three, indicating that the it is beneficial through the ability to address the individual rate requirements of the controllers, which may be at different places in the evolutionary process.  INDIV also displays the best mean fitness ($f$=122562.6, p$<$0.05 compared to STATIC and ADAPT), due to the ability the instantaneous rates.  GLOBAL has a high standard deviation, and so is statistically similar to all other strategies.  This is likely because restarts are driven by global performance only, so individuals may be stuck with suboptimal control in suboptimal rate regions for as long as there is a single population member that is improving the global fitness.  INDIV has the highest low fitness ($f$=117480.6, p$<$0.05 compared to GLOBAL and STATIC), adding further support to the hypothesis that the extra context-sensitivity induced by individually monitoring the controllers for fitness stagnation overcomes the increased disruption to the evolutionary process.  

We note that having suboptimal controllers stuck in suboptimal regions through poor rate setting could potentially improve global performance, if the suboptimal control vectors provide useful genetic code to the global fitness leader.  This depends on the setting of the suboptimal controller vectors and how they interact with the crossover operator, and will be the subject of future research.

We note that both restart strategies have approximately double the mean standard deviation (INDIV=2716, GLOBAL=3376) of ADAPT (1528) and STATIC (1429), showing some of the disruption caused by restarts.  This pattern of high standard deviation for restart strategies is replicated for mean fitness, and low fitness, indiciating that it is a general property of restart strategies.  Disruption is thought be be caused by rates (i) jumping around in the rate space during an experiment, (ii) self-adapting to more promising areas of the rate space, but not before the restart counter is triggered and the rate is reset into an entirely new area, thus disrupting the self-adaptation, and (iii) restarting to a more suboptimal area (effectively wasting the restart).  In the experiment presented here, disruption was evidenced in large standard deviations, rather than direct reductions in fitness and convergence.  As only the fittest final controllers would be used to fly the hexacopter, we conclude that rate restarts are a viable strategy to control evolutionary rate divergence in our ER scenario.

\begin{figure}[t!]
\begin{center}
 \subfloat{\includegraphics[width=8.5cm, height=6.cm]{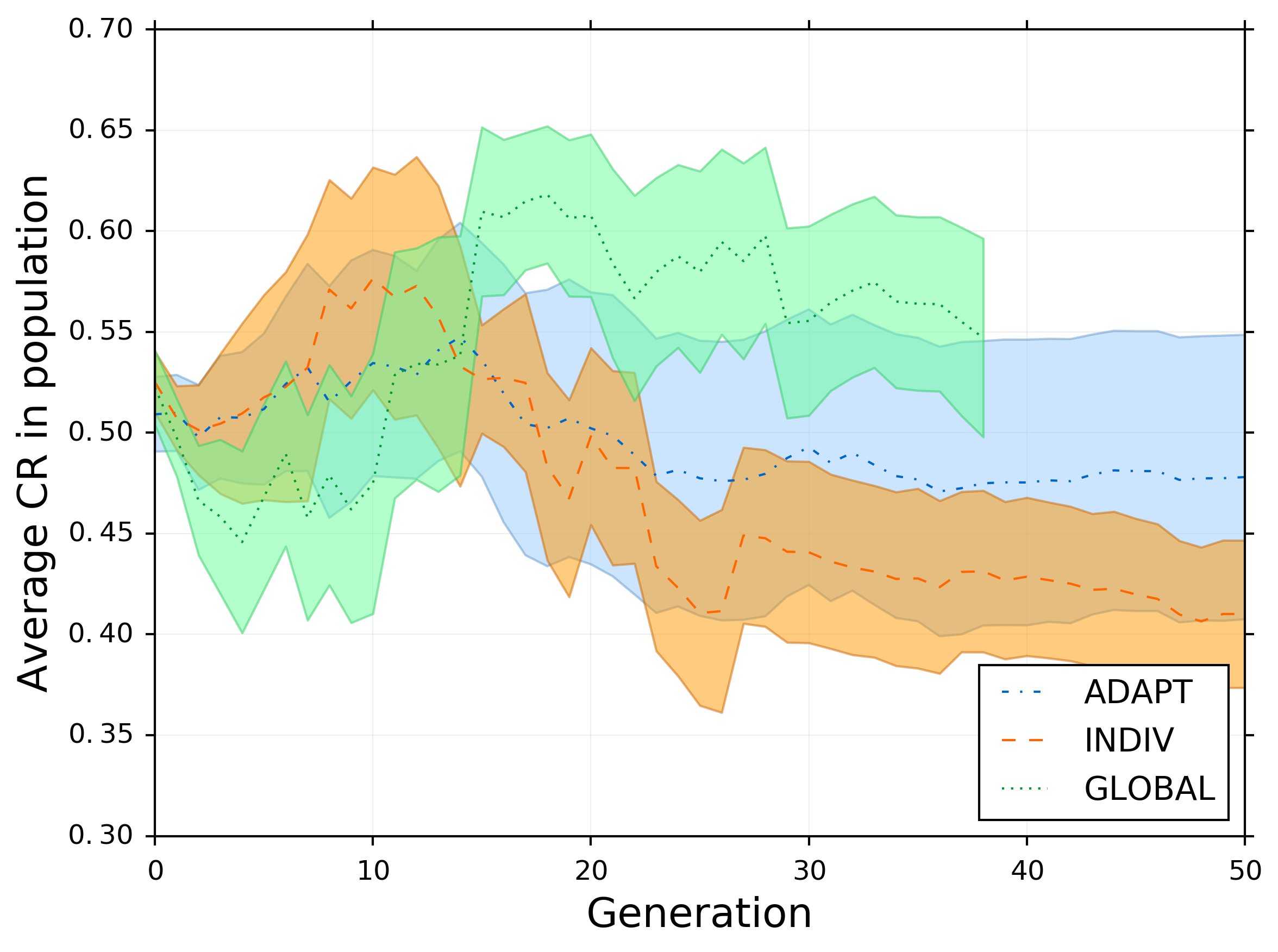} }\\
  \subfloat{\includegraphics[width=8.5cm, height=6.cm]{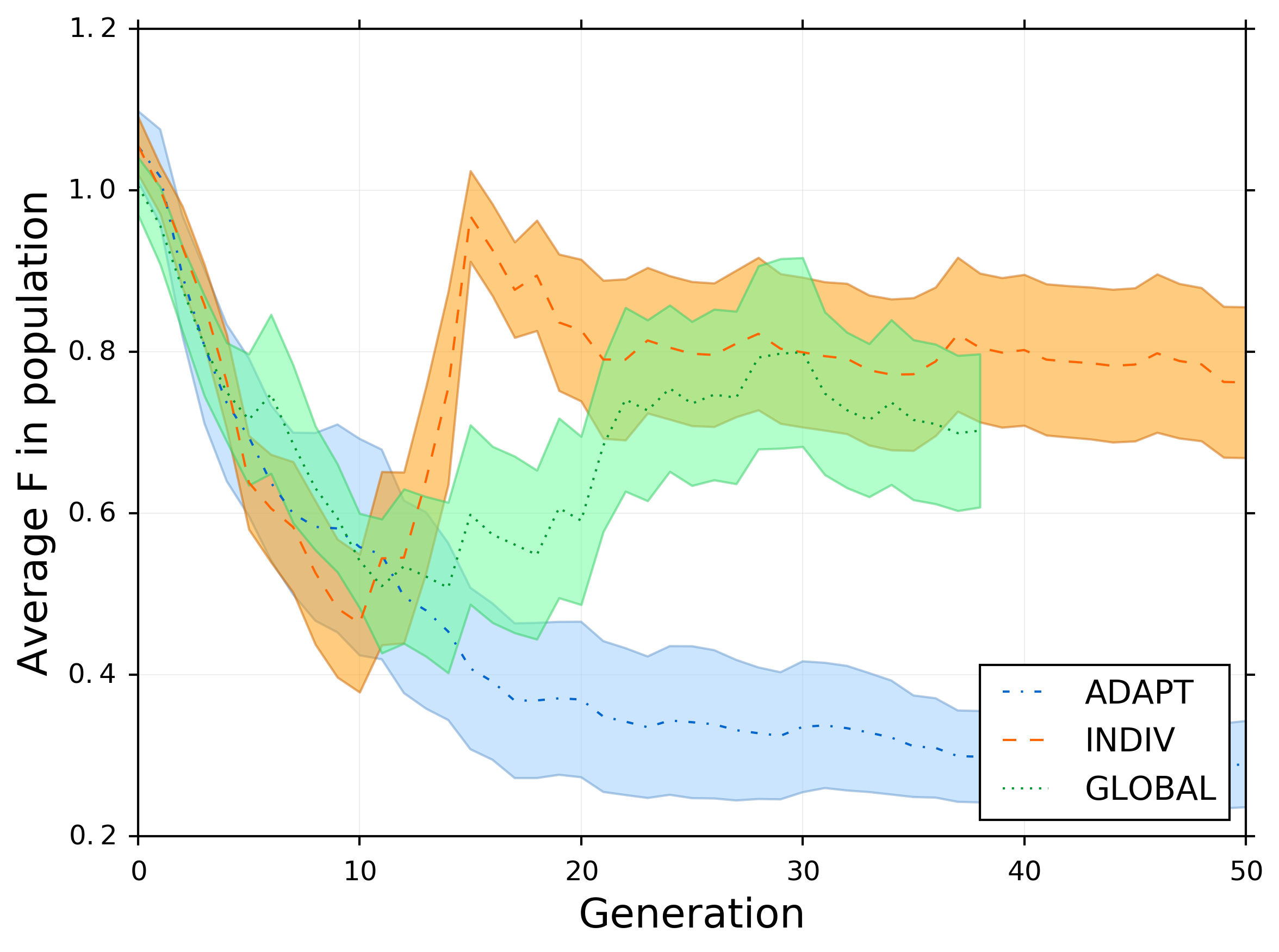} }
\end{center}
\caption[]{Showing mean {\em CR} and {\em F} rates for all self-adaptive experiments throughout the experiments.  Shaded regions denote standard error.}
\label{fff}
\end{figure}

{\bf Rates.} The crossover rate {\em CR} displays no significant differences between the three self-adaptive strategies (INDIV=0.413, GLOBAL=0.536, ADAPT=0.482, Table~\ref{stats}, Fig.~\ref{fff}(a)).  The introduction of restarts significantly increases the mean value of the differential weight {\em F} (INDIV=0.756, GLOBAL=0.758) compared to ADAPT (0.289, both p$<$0.05).  Practically, restarting {\em F} causes reinitialization in the range [0,2] with a mean new value of 1.0 (from\cite{de}), which is subsequently self-adapted down towards the final values.  As ADAPT has no mechanism to quickly alter rates, it converges gradually to its final value, with a corresponding decrease in impact from the donor vector.  When this value becomes too low (Fig.~\ref{fff}(b)), the algorithm struggles to move itself out of local optima.  If these optima do not result in successful controllers, the convergence generation becomes large.  In contrast, the use of restarts in the INDIV and GLOBAL can be seen to increase {\em F} (for INDIV this change is notable after generation 10, for GLOBAL a more gradual increase is observed after generation 13).  The change is more gradual as (i) all rates are reset (meaning resets would settle around the mean value on reinitialisation), and (ii) fewer resets are used as the global fitness stagnation is the trigger.

Restarts occur periodically throughout both INDIV and GLOBAL.  GLOBAL restarts occur mainly between generations 10-30 (Fig.~\ref{fff}), with a mean of 3.7 restarts triggered per experimental repeat, with each restart affecting all 20 individuals for 74 total restarts.  

As INDIV restarts based on individual fitness progression, we note restarts being more uniformly spread across the generations.  As is typical with self-adaptive approaches, the rates themselves vary from generation to generation based on how easily the rates locate successful children.  INDIV uses a mean of 53 restarts per repeat.  This difference is not significant, as the rarity of restart triggering for GLOBAL is offset by each restart affecting the entire population.  The difference in effects of the strategies on parameter evolution is most clearly seen from generations 10-20 in Fig.~\ref{fff}(b).

\section{Conclusions}

In this study, we compared two different implementations for restarting key evolutionary rates during self-adaptive ER experiments, paying particular interest to the level at which restarts are implemented (i.e., individual or population), and compared to two benchmarks, (i) a constant-rate strategy, and a no-restart strategy.  Results indicate that restarts are useful for SA ER experiments, mainly to dissuade premature convergence at local optima caused by poorly-set rates.  These results are in agreement with previous studies using self-adaptive hill-climbing algorithms for ER~\cite{montanier:inria-00566898}, but here the experimentation is expanded to cover population-based algorithms, and consider the two main evolutionary operators, crossover and mutation.

When tested on a hover experiment that optimises PID controllers on a real hexacopter, we note that both INDIV and GLOBAL prevent the extreme outlier convergence generations noted in ADAPT.  Restarts are seen to generate more variance in rate settings.  Disruption is evidenced in the standard deviations in fitness metrics, but not in degredation of attainable fitness values or convergence generations.  Between the two restart strategies, INDIV emerges as our clear preference as it is able to attain higher fitness controllers than all other strategies considered.  

Future research will consider the effect of problem difficulty on the restart strategy performance.  Hover is a relatively simple behaviour, and we envisage restart strategies to have much more of an impact on more challenging tasks, which are more likely to have multimodal fitness landscapes with multiple local optima.  We also wish to experiment with different flying robots, and payloads, and observe the effects of the two on the evolutionary process.

\appendix
\section{Hexacopter State Limits}
\noindent
\begin{math}\displaystyle
l_\mathrm{\omega}: \mbox{max. pitch and roll rate (115$^o$/s)}\\
l_\mathrm{\omega n}: \mbox{pitch and roll rate noise threshold (30$^o$/s)}\\
l_\mathrm{vhn}: \mbox{horizontal velocity noise threshold (5cm/s)}\\
l_\mathrm{vh}: \mbox{max. horizontal velocity (15cm/s)}\\
l_\mathrm{vvn}: \mbox{vertical velocity noise threshold (2cm/s)}\\
l_\mathrm{vv}: \mbox{max. vertical velocity in closed-loop system (20cm/s)}\\
l_\mathrm{a}: \mbox{attitude range limit (15$^o$)}\\
l_\mathrm{h}: \mbox{height range limit (10cm)}\\
l_\mathrm{hc}: \mbox{core height limit (5cm)}\\
l_\mathrm{yc}: \mbox{core yaw limit  (15$^o$)}\\
l_\mathrm{y}: \mbox{yaw range limit  (160$^o$)}\\
l_\mathrm{pc}: \mbox{core position limit (8cm)}\\
l_\mathrm{p}: \mbox{position range limit (20cm)}\\
\end{math}

\section{Fitness Function}

\noindent
\begin{math}\displaystyle
f_\mathrm{cycle}=f_{\mathrm{a}_i}+f_{\mathrm{vh}_i}+f_{\mathrm{vv}_i}+
f_{\mathrm{h}_i}+f_{\mathrm{y}_i}+f_{\mathrm{p}_i}+f_{\mathrm{l}_i}+f_{\mathrm{\omega}_i}
\\
f_\mathrm{l}=
\begin{cases}
0, & \mbox{if PWM limit reached}  \\
1, & \mbox{otherwise}
\end{cases}
\\
f_\mathrm{a}=\mathrm{max}\{1-\frac{|\phi_\mathrm{sp}-\phi|}{l_\mathrm{a}},0\}
+\mathrm{max}\{1-\frac{|\theta_\mathrm{sp}-\theta|}{l_\mathrm{a}},0\} \\
f_\mathrm{vh}=\mathrm{max} \{ 1-\frac{\mathrm{db}\{ 
\sqrt{v_\mathrm{n}^2+v_\mathrm{e}^2}
,l_\mathrm{vhn} \} } {l_\mathrm{vh}}, 0\}
\\
f_\mathrm{vv}=\mathrm{max} \{ 1-\frac{\mathrm{db}\{
|v_\mathrm{v}|,l_\mathrm{vvn} \} } {l_\mathrm{vv}}, 0\}
\\
f_\mathrm{\omega}=
\mathrm{max} \{ 1-\frac{\mathrm{db}\{
|\omega_\mathrm{p}|,l_\mathrm{\omega n} \} } {l_\mathrm{\omega}}, 0\}
+
\mathrm{max} \{ 1-\frac{\mathrm{db}\{
|\omega_\mathrm{q}|,l_\mathrm{\omega n} \} } {l_\mathrm{\omega}}, 0\}
\\
f_\mathrm{h}=
\begin{cases}
\mathrm{max}\{\frac{|h_\mathrm{sp}-h|}{4(l_\mathrm{h}-l_\mathrm{hc})},0\}, & \mbox{if } |h_\mathrm{sp}-h|>l_\mathrm{hc} \\
\frac{3|h_\mathrm{sp}-h|}{4 l_\mathrm{hc}}+\frac{1}{4}, & \mbox{otherwise}
\end{cases}
\\
f_\mathrm{y}=
\begin{cases}
\mathrm{max}\{\frac{|\psi_\mathrm{err}|}{4(l_\mathrm{y}-l_\mathrm{yc})},0\}, & \mbox{if }
|\psi_\mathrm{err}|>l_\mathrm{yc} \\
\frac{3|\psi_\mathrm{err}|}{4 l_\mathrm{yc}}+\frac{1}{4}, & \mbox{otherwise}
\end{cases}
\\
f_\mathrm{p}=
\begin{cases}
\mathrm{max}\{\frac{p_\mathrm{err}}{4(l_\mathrm{p}-l_\mathrm{pc})},0\}, & \mbox{if }
p_\mathrm{err}>l_\mathrm{pc} \\
\frac{3p_\mathrm{err}}{4 l_\mathrm{pc}}+\frac{1}{4}, & \mbox{otherwise}
\end{cases}
\\
p_\mathrm{err}=\sqrt{(p_\mathrm{nsp}-p_\mathrm{n})^2+(p_\mathrm{esp}-p_\mathrm{e})^2}
\\
\psi_\mathrm{err}=\mathrm{wrap}\{\psi_\mathrm{sp}-\psi\}
\\
\mathrm{wrap}\{\alpha\}=\mathrm{atan2}(\sin(\alpha),\cos(\alpha))
\\
\mathrm{db}\{x,l\}=
\begin{cases}
x, & \mbox{if } x>l \\
0, & \mbox{otherwise}
\end{cases}
\vspace{2mm}
\\
f_\mathrm{hover}: \mbox{fitness for hover controller}\\
f_\mathrm{cycle}: \mbox{fitness for one control step}\\
f_\mathrm{a}: \mbox{fitness for pitch and roll}\\
f_\mathrm{vh}: \mbox{fitness for horizontal velocity}\\
f_\mathrm{vv}: \mbox{fitness for vertical velocity}\\
f_\mathrm{h}: \mbox{fitness for height}\\
f_\mathrm{y}: \mbox{fitness for yaw}\\
f_\mathrm{p}: \mbox{fitness for horizontal position}\\
f_\mathrm{\omega}: \mbox{fitness for pitch and roll rates}\\
\end{math}




\bibliographystyle{ACM-Reference-Format}
\bibliography{root} 

\end{document}